\documentclass[10pt,twocolumn,letterpaper]{article}

\usepackage{wacv}              

\usepackage{graphicx}
\usepackage{amsmath}
\usepackage{amssymb}
\usepackage{booktabs}
\usepackage{multirow}
\usepackage{pifont}
\usepackage{xcolor}
\usepackage{stfloats}

\usepackage[pagebackref,breaklinks,colorlinks]{hyperref}

\usepackage[capitalize]{cleveref}
\crefname{section}{Sec.}{Secs.}
\Crefname{section}{Section}{Sections}
\Crefname{table}{Table}{Tables}
\crefname{table}{Tab.}{Tabs.}

\def\wacvPaperID{2844} 
\def\confName{WACV}
\def\confYear{2025}

\begin{document}

\title{DTA: Dual Temporal-channel-wise Attention for Spiking Neural Networks}

\author{Minje Kim\textsuperscript{1}\\
Institution1\\
Institution1 address\\
{\tt\small firstauthor@i1.org}
\and
Minjun Kim\textsuperscript{2}\thanks{Equal contribution.}\\
Institution2\\
First line of institution2 address\\
{\tt\small secondauthor@i2.org}
\and
Xu Yang\textsuperscript{2}\thanks{Corresponding author.} \\
Institution2\\
First line of institution2 address\\
{\tt\small secondauthor@i2.org}
}
\author{
Minje Kim\textsuperscript{1}\thanks{Equal contribution.} \quad 
Minjun Kim\textsuperscript{2}\footnotemark[1]  \quad 
Xu Yang\textsuperscript{2}\thanks{Corresponding author.}\\
    \textsuperscript{1}Promedius Inc.\\
\textsuperscript{2}Beijing Institute of Technology\\
{\tt\small iankimrok@gmail.com, mnjnkai@gmail.com, pyro\_yangxu@bit.edu.cn}
}

\maketitle

\begin{abstract}
   Spiking Neural Networks (SNNs) present a more energy-efficient alternative to Artificial Neural Networks (ANNs) by harnessing spatio-temporal dynamics and event-driven spikes. Effective utilization of temporal information is crucial for SNNs, leading to the exploration of attention mechanisms to enhance this capability. Conventional attention operations either apply identical operation or employ non-identical operations across target dimensions. We identify that these approaches provide distinct perspectives on temporal information. To leverage the strengths of both operations, we propose a novel Dual Temporal-channel-wise Attention (DTA) mechanism that integrates both identical/non-identical attention strategies. To the best of our knowledge, this is the first attempt to concentrate on both the correlation and dependency of temporal-channel using both identical and non-identical attention operations. Experimental results demonstrate that the DTA mechanism achieves state-of-the-art performance on both static datasets (CIFAR10, CIFAR100, ImageNet-1k) and dynamic dataset (CIFAR10-DVS), elevating spike representation and capturing complex temporal-channel relationship. We open-source our code:  \url{https://github.com/MnJnKIM/DTA-SNN}.
\end{abstract}

\vspace{-0.5em}

\section{Introduction}
\label{sec:intro}

\begin{figure*}
  \centering
  \begin{subfigure}{0.72\linewidth}
    \includegraphics[width=0.95\textwidth]{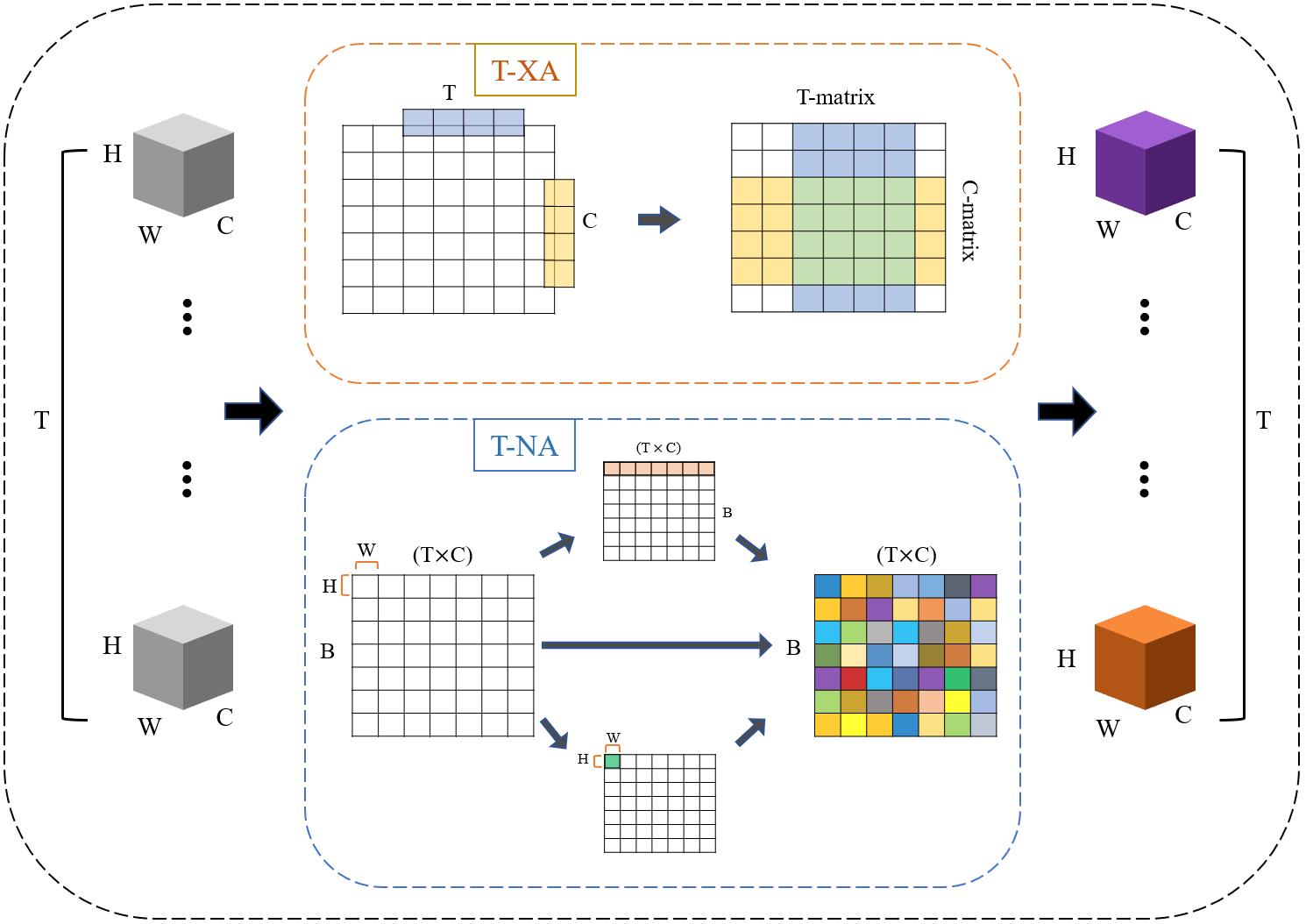}
    \caption{DTA architecture.}
    \label{fig:figure1-a}
  \end{subfigure}
  \hfill
  \begin{subfigure}{0.27\linewidth}
    \includegraphics[width=0.95\textwidth]{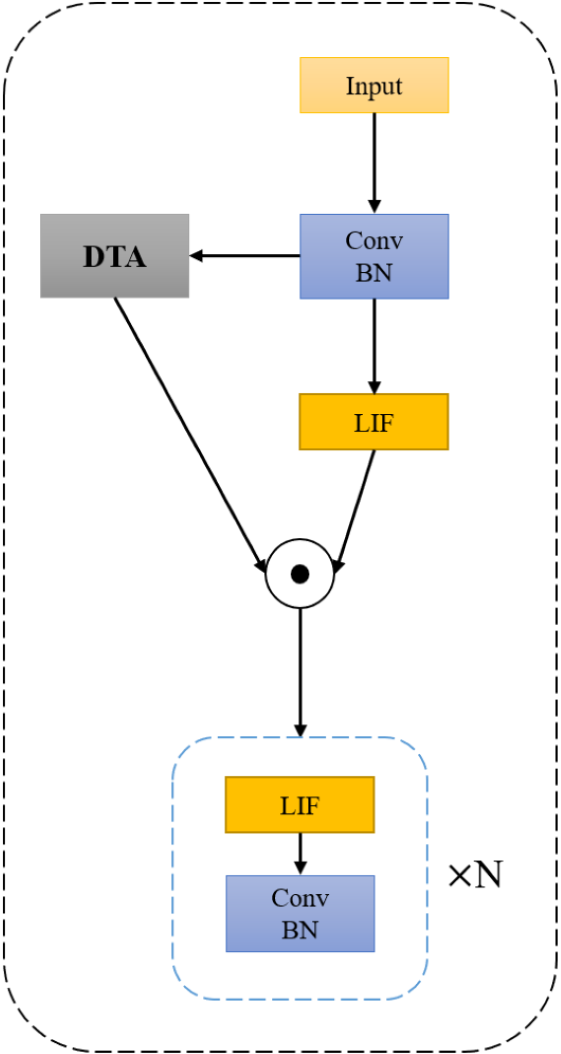}
    \caption{DTA-SNN architecture.}
    \label{fig:figure1-b}
  \end{subfigure}
  \caption{(a) shows the overall workflow of our proposed Dual Temporal-channel-wise Attention (DTA) mechanism. The DTA incorporates Temporal-channel-wise identical Cross Attention (T-XA) and Temporal-channel-wise Non-identical Attention (T-NA) modules to administer an identical operation and non-identical operations across temporal and channel dimensions, respectively. (b) illustrates our proposed DTA-SNN architecture, which embeds a single DTA block into a spiking neural network (SNN).
}
  \label{fig:figure1}
\end{figure*}

Artificial Neural Networks (ANNs) have achieved significant strides in image recognition through attention mechanisms and complex architectures like deeper-wider neural networks. However, these approaches accompany a substantial increase in computational demands and power consumption, which poses challenges for practical applications. To alleviate these challenges, Spiking Neural Networks (SNNs) have emerged, leveraging spatio-temporal dynamics and event-based spikes as activation functions. SNNs replace power-intensive multiply-accumulate operations with more efficient accumulation processes \cite{roy2019towards}. These characteristics enable SNNs to more accurately resemble the actual computations of the human brain \cite{maass1997networks}, offering extreme energy efficiency and low-latency calculations when implemented in neuromorphic hardware \cite{davies2018loihi,merolla2014million,furber2016large}.

Nevertheless, the spike-based information transmission poses significant challenges due to their non-differentiable activation function. To address this issue, the ANN-to-SNN (A2S) conversion method \cite{sengupta2019going} establishes a connection between the neurons of SNNs and ANNs, enabling ANNs to be trained first and then converted to SNNs. While A2S conversion achieves comparable performance to ANNs, it requires numerous time steps and disregards the various alterations of temporal information in SNNs. In contrast, direct training approaches \cite{bohte2000spikeprop,wu2018spatio,neftci2019surrogate}, which train SNNs using surrogate gradient (SG) alleviate the non-differentiable for spike activation. SG-based methods \cite{fang2021deep,li2021differentiable,deng2022temporal} demonstrate effective performance on large datasets with fewer timesteps and can directly process temporal data.

To further advance the capability of SNNs in selectively attending to pertinent information within temporal data, attention mechanism has seen widespread adoption. For instance, TA \cite{yao2021temporal} demonstrates the potential of temporal-wise attention for SNNs. Multi-dimensional Attention methodology \cite{yao2023attention} enhances performance through the non-identical attention across temporal, channel, and spatial dimensions. Additionally, TCJA \cite{zhu2024tcja} effectively extracts spatio-temporal features by combining temporal and channel information with an identical attention at the same stage. Recently, Gated Attention Coding (GAC) \cite{qiu2024gated} leverages non-identical attention with a single multi-dimensional attention block at the input stage.

In this paper, we consider that the attention mechanism depends on the identical and non-identical operation, which yield different expressive capabilities, and both operations improve spike representation. To realize this, we propose a novel dual Temporal-channel-wise attention (DTA) mechanism, which efficiently and effectively enhances feature representation for SNNs by combining the benefits of both identical and non-identical attention operations, leading to seizure of complex temporal-channel dependency. As illustrated in \Cref{fig:figure1-a}, the DTA mechanism is composed of two independent attention strategies: Temporal-channel-wise identical Cross Attention (T-XA) and Temporal-channel-wise Non-identical Attention (T-NA). Initially, the T-XA module performs identical operation of temporal and channel dimensions to ensure elaborate temporal-channel correlation via cross attention. In contrast, the T-NA module addresses non-identical operations to interpret  both local and global dependencies of the temporal-channel, leading to abundant spike representation. Additionally, as visualized in \Cref{fig:figure1-b}, we implement SNNs consisting of a single DTA block by adopting MS-ResNet structure \cite{hu2024advancing} and GAC scheme \cite{qiu2024gated} to prove that a single DTA block can outperform previous studies using multiple attention blocks in SNNs. Our implementation alleviates the high computational cost, memory usage, and low interpret-ability associated with multiple attention blocks. Our contributions are summarized below:
\begin{itemize}
\item To the best of our knowledge, this is the first attempt to incorporate both identical/non-identical attention mechanisms into the SNNs. Our proposed DTA mechanism notes that the expressive capabilities of attention relies on a dual temporal-channel-wise perspective.
\item We introduce a novel T-XA module, which simultaneously considers the correlation of temporal and channel information with the identical attention operation. We also show that the T-NA module employs non-identical operations to handle the combined temporal-channel dimension from both intra- and inter-dependencies.
\item Our proposed DTA mechanism, incorporating a single DTA block, achieves state-of-the-art performance on static and dynamic datasets. Experimental results demonstrate exceptional accuracy, with 96.73{\%} /81.16{\%} on CIFAR10/100 \cite{CIFAR10}, 71.29{\%} on ImageNet-1k\cite{krizhevsky2017imagenet}, and 81.3{\%} on CIFAR10-DVS\cite{li2017cifar10}.\end{itemize}


\begin{figure*}
  \centering
  \begin{subfigure}{0.21\linewidth}
    \includegraphics[width=\textwidth]{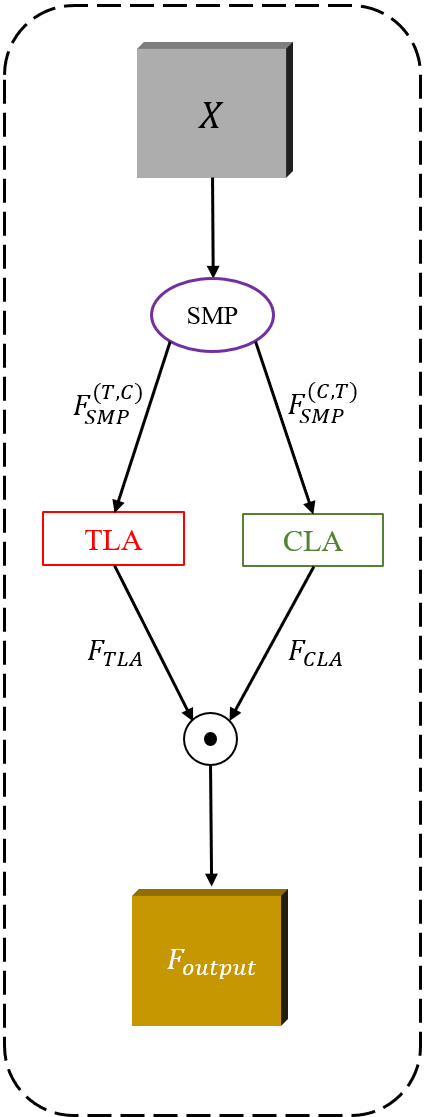}
    \caption{T-XA module.}
    \label{fig:figure2-a}
  \end{subfigure}
  \hfill
  \begin{subfigure}{0.21\linewidth}
    \includegraphics[width=\textwidth]{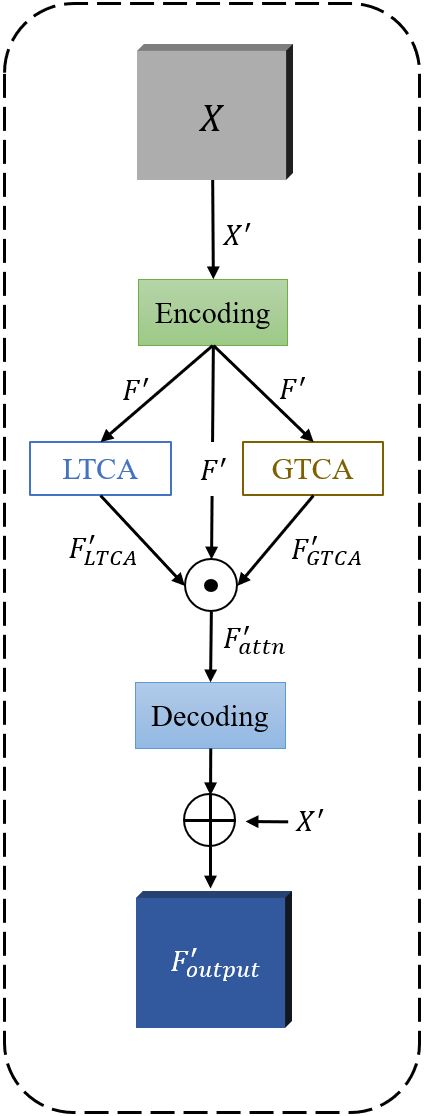}
    \caption{T-NA module.}
    \label{fig:figure2-b}
  \end{subfigure}
  \hfill
  \begin{subfigure}{0.56\linewidth}
    \includegraphics[width=\textwidth]{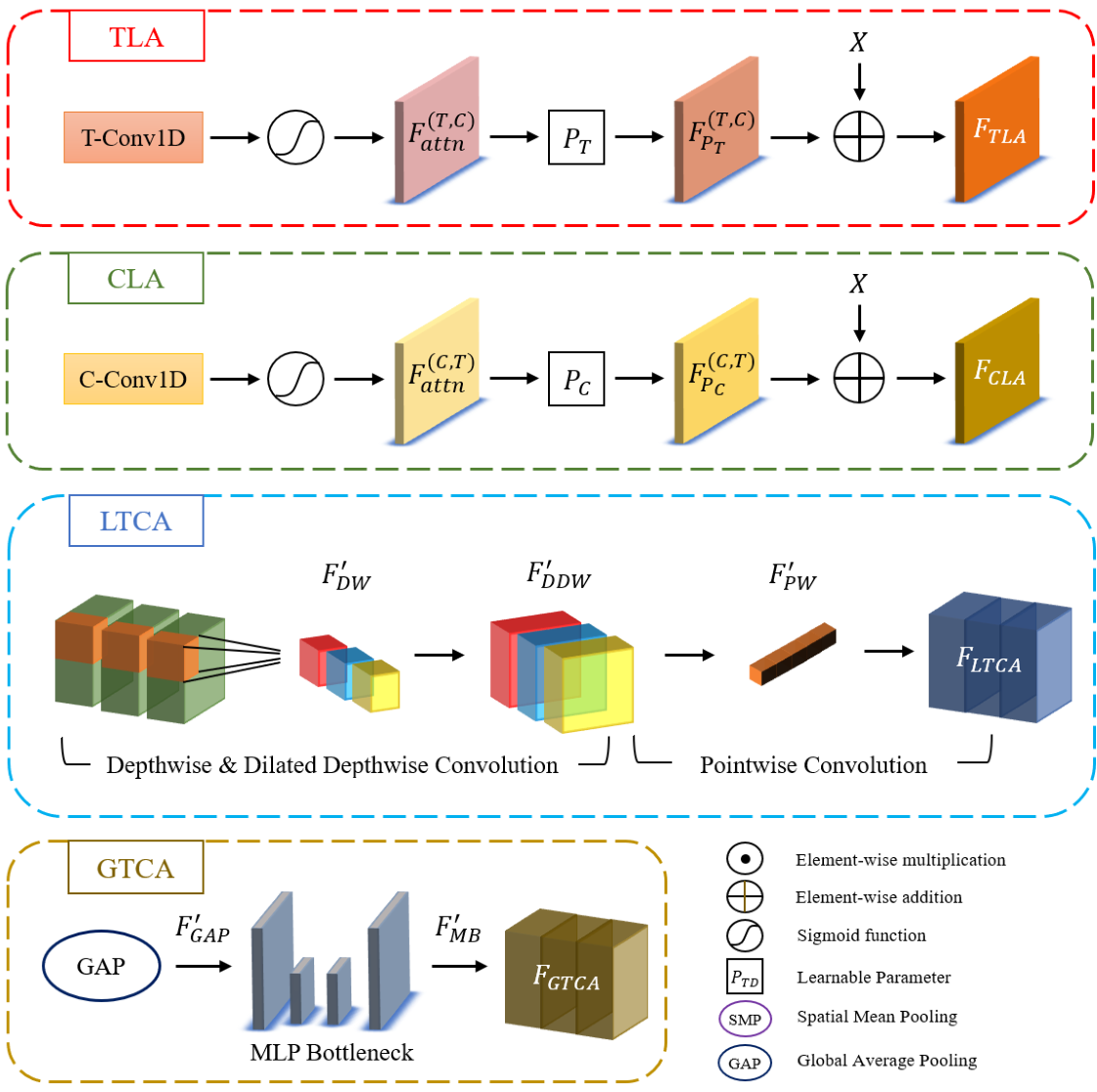}
    \caption{Details of T-XA and T-NA modules.}
    \label{fig:figure2-c}
  \end{subfigure}
  \caption[c]{(a) presents the overall structure of the T-XA module, (b) illustrates the design of the T-NA module, and (c) shows the overall architecture of the Temporal Local Attention (TLA), the Channel Local Attention (CLA), the Local Temporal-Channel Attention (LTCA), and the Global Temporal-Channel Attention (GTCA).
}
  \label{fig:figure1}
\end{figure*}

\section{Related Works}
\label{sec:RW}


\subsection{SNN Training Methods}
The two primary training approaches for SNNs are the A2S conversion and the DT of SNNs. The A2S conversion methods typically involve transforming pre-trained ANNs into SNNs by replacing the activation functions with spiking neurons. For instance, these conversion methods have been proposed, including threshold balancing \cite{diehl2015fast}, spiking equivalents \cite{rueckauer2017conversion}, soft reset \cite{han2020rmp}, calibration algorithms \cite{li2021free,hao2023bridging}, rate norm layer \cite{ding2021optimal}, quantization \cite{li2022quantization, hu2023fast, wang2023toward}, and residual membrane potential \cite{wang2023toward,hao2023reducing}. While these methods are predominantly applied to CNNs, recent works \cite{spikeziptf2024,jiang2024spatiotemporal} have explored their application to Transformers. However, A2S conversion methods have not yet surpassed the performance of ANNs, and the rate coding scheme of these methods constrains the temporal dynamics of SNNs, leading to elevated energy consumption.

The DT approaches \cite{fang2021deep,neftci2019surrogate} train SNNs from scratch and utilize SG to resolve non-differentiability of spiking neurons. These methods are capable of handling temporal data, such as event-based datasets, with fewer time-steps required for training. Conventional spiking neuron models, including integrate-and-fire (IF) and leaky integrate-and-fire (LIF) \cite{wu2018spatio}, which implements spike-based information transmission, while recent advancements introduced novel models such as PLIF \cite{Fang_2021_ICCV}, GLIF \cite{yao2022glif}, and CLIF \cite{huang2024clif}. Early DT methods, like BPTT \cite{neftci2019surrogate}, iteratively update gradients through spatial and temporal dimensions, and  \cite{wu2019direct,guo2022loss} demonstrated efficiency on both dynamic and static datasets. However, these methods adopted SG as a tanh-like function, causing explosion or vanishing of gradient. Recent researches addressed these challenges by proposing various SG methods, including triangular shapes \cite{li2021differentiable,deng2022temporal,rathi2020diet}, sigmoid functions \cite{zenke2021remarkable,wozniak2020deep}, and adaptive-learnable SG \cite{wang2023adaptive,lian2023learnable,deng2023surrogate}. Additionally, several studies \cite{zheng2021going,duan2022temporal,jiang2024tab} explored temporal-based batch normalization to manage temporal covariate shift.
\subsection{Attention Mechanism in SNNs}
The attention mechanism in CNNs overcomes inherent limitations by focusing on salient regions or channels within an image and selectively emphasizing prominent features. This advantage has catalyzed research into extending these techniques to SNNs, investigating how attention mechanisms can be harnessed to enhance performance of SNNs. TA \cite{yao2021temporal} presented a temporal-wise attention mechanism with the squeeze-and-excitation block, which assigns attention factors to each temporal-wise input frame in SNNs, and demonstrated the potential of attention mechanisms in processing temporal data.  The multi-dimensional attention methodology \cite{yao2023attention} promotes membrane potentials to use attention weights via sequential application of temporal/channel/spatial-wise attention, and their performance gains were validated across various benchmark datasets. TCJA \cite{zhu2024tcja} has proven the effectiveness of joint attention by integrating temporal and channel information with 1-D convolution, enabling the low-cost extraction of spatio-temporal features in SNNs. Nonetheless, the aforementioned methods rely on direct coding, which necessitates the use of multiple attention blocks. Consequently, these methods frequently generate iterative-similar outputs at each time step, resulting in weak spike representation and limited spatio-temporal dynamics. To address this issue, Gated Attention Coding (GAC) \cite{qiu2024gated} effectively captured variations in temporal information using a single attention block and demonstrated strong performance on static datasets.

\section{Methods}
\label{sec:Methods}
\subsection{Neuron Model and Surrogate Gradient for SNNs}
We adopt the iterative LIF neuron model proposed in \cite{wu2018spatio}, and its dynamics can be described as follows:
\begin{equation}
  u^n(t) = \tau u^n(t-1) \odot (1-s^n(t-1)) + c^n(t),
  \label{eq:important}
\end{equation}
where $u^n(t)$ represents the membrane potential of spiking neuron in the $n$-th layer at the time step $t$, $\tau$ is the time constant determining the decay rate of membrane potential, and $(1 - s^n(t-1))$ resets the potential to 0 after a spike. The symbol $\odot$ denotes element-wise multiplication.

The synaptic input current $c^n(t)$ is computed as:
\begin{equation}
  c^n(t) = W^n \star s^{n-1}(t),
  \label{eq:also-important}
\end{equation}
where $W^n$ is the synaptic weight matrix, and $s^{n-1}(t)$ represents the output spikes from the previous layer at the time step $t$. The symbol $\star$ denotes the synaptic operation, which involves either convolutional or fully connected layers.

When the membrane potential $u^n(t)$ exceeds the threshold, a spiking output $s^n(t)$ is generated. It is defined as follows:
\begin{equation}
  s^n(t) = \mathbb{H}(u^n(t)-V_{th}),
  \label{eq:also-important}
\end{equation}
where $s^n(t)$ is the binary spiking output generated by the Heaviside step function $\mathbb{H}$, and $V_{th}$ is the firing threshold. The Heaviside step function yields 1 when $(u^n(t)-V_{th})$ is non-negative, and 0 otherwise.
Additionally, we update the weight of SNNs using the spatial-temporal backpropagation\cite{wu2018spatio}:
\begin{equation}
  \Delta W^n = \sum_{t} \frac{\partial L}{\partial W^n} = \sum_{t} \frac{\partial L}{s^n(t)}\frac{\partial s^n(t)}{\partial u^n(t)} \frac{\partial u^n(t)}{\partial c^n(t)}\frac{\partial c^n(t)}{\partial W^n},
  \label{eq:also-important}
\end{equation}
From the above equation, the term $\partial s^n(t)/\partial u^n(t)$ indicates gradient of the spiking function, which remains zero value with the exception of the scenario where $u^n(t)$ equals to $Vth$. To circumvent the non-differentiability, previous studies have introduced diverse forms of SG methods \cite{deng2022temporal,fang2021deep,li2021differentiable}. We implement the triangular SG method, as follows:

\begin{equation}
  \frac{\partial s^n(t)}{\partial u^n(t)} = \begin{cases} \alpha (1 -  \alpha \cdot {\delta}^n(t)) & \text{if} \quad {\delta}^n(t)< \frac{1}{\alpha} 
  
  \\ \text{\hspace{3em}0} & \text{otherwise} \end{cases}
,
  \label{eq:also-important}
\end{equation}
where the constant $\alpha$ determines the maximum activation range of the gradient by adjusting the non-zero interval, and ${\delta}^n(t)$ denotes $\vert u^n(t)-V_{th}\vert $.
\subsection{Dual Temporal-channel-wise Attention}

We present the Dual Temporal-channel-wise Attention (DTA) mechanism, which integrates two identical/non-identical attention strategies: Temporal-channel-wise identical Cross Attention (T-XA) and Temporal-channel-wise Non-identical Attention (T-NA). Our proposed DTA mechanism aims to efficiently and richly enhance feature representation by combining the benefits of both identical and non-identical attention, effectively capturing complex temporal-channel correlation and temporal-channel dependency:
\begin{equation}
  O_{DTA} = 
   \sigma(O_{T-XA} \odot O_{T-NA}) \odot Spikes,
  \label{eq:also-important}
\end{equation}
The outputs of the DTA block are derived by combining the outputs of T-XA and T-NA through element-wise multiplication $\odot$ with the sigmoid function $\sigma$. This process leverages the strengths of both identical and non-identical attention mechanisms, by simultaneously integrating the emphasis with spikes. Specifically, as shown in \Cref{fig:figure1-a}, the T-XA module executes an identical operation across both temporal and channel dimensions, ensuring fine-grained temporal-channel correlation. Contrastively, the T-NA module addresses inputs by applying non-identical operations that comprehend both local and global dependencies across the temporal and channel dimensions, enabling abundant feature representation in SNNs. 
\subsection{Temporal-channel-wise identical Cross Attention}

We first use the T-XA module to refine XA in the DTA block, as shown in \Cref{fig:figure2-a}. The T-XA module is a global attention branch, which consists of Temporal-wise Local Attention ($TLA(\cdot)$) and Channel-wise Local Attention ($CLA(\cdot)$) branches. To implement the T-XA module, we consider the correlation of temporal and channel information through cross attention and the operation of the T-XA module, as follows: 
\begin{equation}
  T{-}XA(X) = TLA(X)\odot CLA(X),
  \label{eq:also-important}
\end{equation}
where $X\in \mathbb{R}^{(T \times C \times H \times W)}$ is the inputs from pre-synaptic neurons. We transform the $X$ into enhanced features $F_{SMP}^{(TD, SD)} \in \mathbb{R}^{(TD \times SD)}$. Here, $TD$ and $SD$ represent that the target and subtarget dimensions, respectively, and we primarily focus on $TD$, considering $SD$ in the each local branch. Additionally, we use Spatial Mean Pooling (SMP) to capture the global cross-acceptance field, which reflects interactive correlation, supplying efficiently extracted features. The $F_{SMP}^{(TD, SD)}$ is generated through SMP over the spatial dimensions $H$ and $W$ of the $X$, and the $F_{SMP}^{(TD, SD)}$ is defined as:
\begin{align}
  F_{SMP}^{(TD, SD)} = SMP(X, TD, SD),\nonumber \\
 =\frac{1}{H \times W} \sum_{i=1}^H \sum_{j=1}^W X^{(TD, SD)}.
  \label{eq:also-important}
\end{align}

Each local attention branch simultaneously processes the temporal and channel dimensions while using a small number of optimized parameters to provide attention, thereby enriching the temporal-channel representation in SNNs. As shown in \Cref{fig:figure2-c}, We obtain the features $F_{LA}$ from the local attention branch and the definition is as follows: 
\begin{equation}
  F_{LA} = P_{TD}(\sigma (conv(F_{SMP}^{(TD, SD)}))) \oplus X^{(TD, SD)},
  \label{eq:also-important}
\end{equation}
where the $conv(\cdot)$ is 1D convolution operation along the $TD$ with the $SD$, the feature map $F^{TD}$ obtained by this operation is generated through the sigmoid function $\sigma$, and the attention map $F^{TD}_{attn}$ is scaled with the learnable parameter $P_{TD}$. Finally, $F^{TD}_{attn}$ and $F^{(TD, SD)}$ are combined through the residual operation $\oplus$ to obtain $F_{LA}$. $F_{TLA}$ and $F_{CLA}$, obtained from $TLA(\cdot)$ and $CLA(\cdot)$ respectively, are then merged using the element-wise multiplication operation to output features $F_{output}$ of T-XA.
\subsection{Temporal-channel-wise Non-identical Attention}

As illustrated in \Cref{fig:figure2-b}, we introduce the T-NA module to effectively address both intra/inter-dependencies between temporal and channel information via non-identical operations. Concretely, our T-NA module consists of local and global temporal-channel attention, and given inputs $X\in \mathbb{R}^{T\times C\times H\times W}$ are reshaped inputs $X’\in \mathbb{R}^{(T\times C)\times H\times W}$ for alleviating computational cost of attention such as 3D convolution or 3D pooling. The $X'$ is projected to attain features $F'$ via encoding operation for the enriched representation of $X'$: 
\begin{equation}
  F' = f(X'), 
  \label{eq:also-important}
\end{equation}
where functional operation $f(\cdot)$ includes 2D convolution operation with $1 \times 1$ kernel and GELU function. 

As shown in \Cref{fig:figure2-c}, to address intra-dependency of temporal-channel information, we propose the local temporal-channel attention (LTCA) mechanism, which processes the $F’$ using several convolution operations as follows:
\begin{equation}
  LTCA(F') = f_{PW}( f_{DDW}(f_{DW}(X') )),
  \label{eq:also-important}
\end{equation}
where $f_{PW}$, $f_{DDW}$ and $f_{DW}$ are point-wise, dilation-depth-wise and depth-wise convolution operations, respectively, and those convolution operations effectively capture intra-dependency of the temporal-channel while maintaining computational efficiency. Concurrently, we present the global temporal-channel attention (GTCA) mechanism to enable adaptive responses to inter-information of the temporal-channel, as follows:
\begin{equation}
  GTCA(F’) = f_{MB}(f_{GAP}(F’)),
  \label{eq:also-important}
\end{equation}
where $f_{GAP}(\cdot)$ indicates global average pooling, which compresses $F'$ by averaging spatial dimensions to highlight the global context of the temporal-channel features. Then, the emphasized features are manipulated using the MLP bottleneck structure $f_{MB}(\cdot)$, which consists of the $Linear{-}ReLU{-}Linear$ sequence. This $f_{MB}$ first squeezes and then expands them in order to reinforce the feature representation by effectively incorporating inter-dependency of temporal-channel information. Subsequently, the attention maps from $LTCA(\cdot)$ and $GTCA(\cdot)$ are integrated through element-wise multiplication with the original $F'$ to yield the attention features $F’_{attn}$:
\begin{equation}
  F’_{attn} = F'_{LTCA} \odot F’_{GTCA} \odot F'.
  \label{eq:also-important}
\end{equation}

Finally, we exploit decoding operation $f_{conv}(\cdot)$ with $1 \times 1$ convolution to handle the $F’_{attn}$, and utilize residual connection for facilitating more accurate attention and contributing more stable training in the T-NA module:
\begin{equation}
  F’_{output} = f_{conv}(F'_{attn}) \oplus X'.
  \label{eq:also-important}
\end{equation}
\begin{table}[h]
  \centering
  {\small{
  \begin{tabular}{@{}lcccc@{}}
    \toprule
    Layer & $C_{Out}$ & $I_{Out}$ & ResNet-18 & ResNet-34 \\ \midrule
    Conv1 & 32$\times$32 & 112$\times$112     & 3$\times$3, 64, s=1  & 7$\times$7, 64, s=2 \\ \midrule
    Conv2 & 32$\times$32 & 56$\times$56      & \begin{tabular}[c]{@{}l@{}}{[}3$\times$3, 128{]}\\ {[}3$\times$3, 128{]}\end{tabular} \hspace{0.3em} \makebox[0pt][c]{$\times3$} & \begin{tabular}[c]{@{}l@{}}{[}3$\times$3, 64{]}\\ {[}3$\times$3, 64{]}\end{tabular} \hspace{0.3em} \makebox[0pt][c]{$\times3$} \\ \midrule
    Conv3 & 16$\times$16 & 28$\times$28      & \begin{tabular}[c]{@{}l@{}}{[}3$\times$3, 256{]}\\ {[}3$\times$3, 256{]}\end{tabular} \hspace{0.3em} \makebox[0pt][c]{$\times3$} & \begin{tabular}[c]{@{}l@{}}{[}3$\times$3, 128{]}\\ {[}3$\times$3, 128{]}\end{tabular} \hspace{0.3em} \makebox[0pt][c]{$\times$4} \\ \midrule
    Conv4 & 8$\times$8 & 14$\times$14      & \begin{tabular}[c]{@{}l@{}}{[}3$\times$3, 512{]}\\ {[}3$\times$3, 512{]}\end{tabular} \hspace{0.3em} \makebox[0pt][c]{$\times$2} & \begin{tabular}[c]{@{}l@{}}{[}3$\times$3, 256{]}\\ {[}3$\times$3, 256{]}\end{tabular} \hspace{0.3em} \makebox[0pt][c]{$\times$6} \\ \midrule
    Conv5 & -  & 7$\times$7       & - & \begin{tabular}[c]{@{}l@{}}{[}3$\times$3, 512{]}\\ {[}3$\times$3, 512{]}\end{tabular} \hspace{0.3em} \makebox[0pt][c]{$\times$3} \\ \midrule
         & \multicolumn{2}{c}{AveragePooling,}       & FC-10/100 & FC-1000\\ \bottomrule
  \end{tabular}
  }}
  \caption{Architecture of MS-ResNet utilized for CIFAR10/100 and ImageNet-1k datasets in DTA-SNNs. $C_{Out}$ and $I_{Out}$ represent the sizes of output features in the CIFAR10/100 and ImageNet-1k datasets, respectively.}
  \label{tab:table1}
\end{table}

\vspace{-0.66em}

\section{Experiments}
\label{sec:Experiments}
In \Cref{subsec:settings}, we first describe the experimental setup used to evaluate our proposed DTA mechanism for SNNs. Next, in \Cref{subsec:comparative}, we demonstrate the effectiveness of the DTA mechanism through a comparative analysis with state-of-the-art (SOTA) methods. Finally, we conduct an ablation study to further investigate the impact of the individual modules of the DTA mechanism, as detailed in \Cref{subsec:ablation}.

\subsection{Settings} \label{subsec:settings} \noindent\textbf{Datasets.} We evaluate our DTA mechanism on four types of classification benchmark datasets, encompassing both static and dynamic datasets. First, we use CIFAR10 and CIFAR100 \cite{CIFAR10}, widely recognized static image classification benchmarks with 10 and 100 classes of natural images, respectively. Next, we employ ImageNet-1k \cite{krizhevsky2017imagenet}, a large-scale static image classification dataset comprising 1,000 classes of diverse objects. Finally, we utilize CIFAR10-DVS \cite{li2017cifar10}, an event-based image classification dataset derived from scanning each sample of the static CIFAR10 dataset using DVS cameras.

\begin{table}[h]
  \centering
    \setlength{\tabcolsep}{4pt}
  {\small{
  \begin{tabular}{@{}lllllll@{}}  
    \toprule
    Dataset     & \parbox{1cm}{Batch \\ Size} & Epochs & \parbox{1cm}{Time Step} & \parbox{1cm}{Initial LR} & Decay \\
    \midrule
    CIFAR10     & 64         & 250    & 4/6       & 0.1      & 5e-5  \\
    CIFAR100    & 64         & 250    & 4/6       & 0.1      & 5e-5  \\
    ImageNet-1k    & 128        & 200    & 4/6       & 0.1      & 1e-5  \\
    CIFAR10-DVS & 64       & 1000   & 10        & 0.05      & 0     
    \\
    \bottomrule
  \end{tabular}
  }}
  \caption{Hyper-parameter training settings for DTA-SNNs.}
  \label{tab:table2}
\end{table}

\begin{table*}[bp]
    \centering
    {\small{
    \begin{tabular}{@{}ccccccc@{}}
    \toprule
    Method    & Type           & NN Architecture & Parameter(M)  & Time Step                                     & CIFAR10 Acc(\%) & CIFAR100 Acc(\%)                                                     \\ \midrule
                               RMP-SNN \cite{han2020rmp}    & A2S & VGG-16 & 33.64/34.01 & 2048    & 93.63  &70.93                                                           \\
                               SRP \cite{hao2023reducing}       & A2S         & VGG-16      & 33.64/34.01         & 32/64  &  95.42/95.40 & 77.01/77.10                                                               \\
                               QCFS \cite{ding2021optimal}      & A2S         & VGG-16/ResNet-18      & 33.64/11.22         & 32/64  & 95.54/95.55 & 76.45/76.37                                                               \\
                               Diet-SNN \cite{rathi2020diet}  & DT & ResNet-20  & -  & 10      & 92.54 & 64.07                                                            \\
                               Dspike \cite{li2021differentiable}    & DT & ResNet-18 & 11.17/11.22 & 6   & 94.25 & 74.24                            \\
                               STBP-tdBN \cite{zheng2021going}    & DT & ResNet-19 & 12.63 & 4/6  & 92.92/93.16 &-                                          \\
                               TET \cite{deng2022temporal}      & DT & ResNet-19 & 12.63/12.67 & 4/6  & 94.44/94.50 & 74.47/74.72                                        \\
                               TEBN \cite{duan2022temporal}       & DT & ResNet-19 & 12.63/12.67 & 4/6  & 95.58/95.60 & 76.13/76.41                                 \\
                               GLIF \cite{yao2022glif}      & DT & ResNet-19 & 12.63/12.67 & 4/6  & 94.85/95.03 & 77.05/77.35                                             \\ 
                               Real Spike \cite{guo2022real} & DT & ResNet-19/VGG-16 & 12.63/- & 4/6  & 95.60/95.71 & 70.62/71.17                                  \\
                               Ternary Spike \cite{guo2024ternary} & DT & ResNet-19 & 12.63 & 1/2  & 95.28/95.60 & 78.13/79.66                                  \\
                               TAB \cite{jiang2024tab}& DT & ResNet-19 & 12.63/12.67 & 4/6   & 95.94/96.09 & 76.81/76.82                                              \\
                               CLIF \cite{huang2024clif}     & DT & ResNet-18 & 11.17/11.22 & 4/6 & 96.01/96.45 & 79.69/80.58                           \\
                               MPBN \cite{guo2023membrane}& DT & ResNet-19 & 12.63/12.67 & 1/2  & 96.06/96.47 & 78.71/79.51                                           \\ 
                               GAC-SNN \cite{qiu2024gated}      & DT             & MS-ResNet-18   & 12.63/12.67   & 4/6 & 96.24/96.46 & 79.83/80.45  \\  \midrule
                                \textbf{DTA-SNN(Ours)}       & DT             & MS-ResNet-18  & 12.99/13.03    & \begin{tabular}[c]{@{}l@{}}4\\ 6\end{tabular} & \begin{tabular}[c]{@{}l@{}}\textbf{96.50 $\pm$ 0.09}\\ \textbf{96.73 $\pm$ 0.11}\end{tabular} & \begin{tabular}[c]{@{}l@{}}\textbf{79.94 $\pm$ 0.08}\\ \textbf{81.16 $\pm$ 0.27}\end{tabular}  \\ \bottomrule                                            
    \end{tabular}
    }}
    \caption{Comparison with SOTA on CIFAR10/100.}
    \label{tab:table3}
\end{table*}

\noindent\textbf{Implementation Details.} We implement our proposed DTA mechanism using the PyTorch framework, conduct experiments on several NVIDIA A100 GPUs, and adopt the MS-ResNet structure \cite{hu2024advancing} to build SNNs that consist of a single DTA block, as visualized in \Cref{fig:figure1-b}. For experiments on the CIFAR10/100, we utilize MS-ResNet18, whereas MS-ResNet34 is employed for the ImageNet-1k dataset, as shown in \Cref{tab:table1}. In the case of the CIFAR10-DVS dataset, the input size is set to 48×48, which leads to adjustments in the internal output sizes of the MS-ResNet18 architecture. Furthermore, we use the SGD optimizer with 0.9 momentum and a cosine annealing schedule \cite{loshchilov2016sgdr} for all our experiments, with detailed hyper-parameters described in \Cref{tab:table2}. 
Additionally, the hyper-parameters for the iterative LIF neuron are configured with the firing threshold $V_{th}$ of 1.0 and the time constant $\tau$ of 0.5. For the overall experiments, we follow widespread data augmentation strategies from previous studies \cite{qiu2024gated,yao2023attention,duan2022temporal, huang2024clif,guo2023membrane,jiang2024tab,zhou2023spikformer,zhou2023spikingformerspikedrivenresiduallearning,yao2023spikedriven}. Notably, in contrast to other SOTA algorithms\cite{zhu2024tcja, huang2024clif, fang2024parallel} that use TET loss \cite{deng2022temporal}, we demonstrate the effectiveness of our attention mechanism using standard cross-entropy loss.

\subsection{Comparisons with SOTA methods} \label{subsec:comparative}
\noindent\textbf{CIFAR10/100}. On the CIFAR10/100 datasets, each experiment was conducted three times, with the mean accuracy and standard deviation reported and summarized in \Cref{tab:table3}. Our proposed method outperformed previous best results with higher accuracy and fewer time steps. Specifically, DTA-SNN achieves SOTA performance with  96.73{\%} top-1 accuracy at 6 time steps and 96.50{\%} accuracy at 4 time steps. Compared to non-identical attention method, DTA-SNN surpassed the GAC-SNN accuracy of 96.46{\%} at 6 time steps with an accuracy of 96.50{\%} at just 4 time steps on the CIFAR10 dataset. Moreover, we also obtained 0.71{\%} higher performance than GAC-SNN at 6 time steps on the CIFAR100 dataset. These results demonstrate the effectiveness of our dual attention operations. 

\begin{table*}[hbt!]
    \centering
    {\small{
    \begin{tabular}{@{}lcccccc@{}}
    \toprule
    &Method    & Type            & NN Architecture  &Parameter(M) & Time Step                                     & Accuracy(\%)                                                      \\ \midrule
     & SRP \cite{hao2023reducing}      & A2S        & Modified-VGG-16   & 138.36 & 32/64  & 69.35/69.43              \\
     & QCFS \cite{ding2021optimal}      & A2S         & ResNet-34      & 21.79         &  32                                          &  69.37 \\
     
     & RMP-SNN \cite{han2020rmp}    & A2S & ResNet-34 & 21.79 & 4096                                             & 69.89 \\
    & Diet-SNN \cite{rathi2020diet}  & DT & VGG-16  & -  & 5                                            & 69.00                                                             \\
    & Dspike \cite{li2021differentiable}   & DT & ResNet-34  & 21.79 &  6                                             & 68.19                                                             \\
    & TET \cite{deng2022temporal}       & DT & ResNet-34 & 21.79 & 6                                             & 64.79                                                             \\
    & SEW-ResNet \cite{fang2021deep} &DT &SEW-ResNet-34 & 21.79 & 4 & 67.04 \\
    & MS-ResNet \cite{hu2024advancing} & DT & Ms-ResNet-34   & 21.80   & 6                                             & 69.43                                                    \\
    & GLIF \cite{yao2022glif}      & DT & ResNet-34  & 21.79  & 6                                             & 69.09                                                             \\ 
    & Real Spike \cite{guo2022real} & DT & ResNet-34 & 21.79  & 4                                             & 67.69                                                            \\
    & Ternary Spike \cite{guo2024ternary} & DT & ResNet-34 & 21.79 & 4  & 70.12                                 \\
    & TAB \cite{jiang2024tab}& DT & ResNet-34 & 21.79 & 4                                             & 67.78                                                            \\
    & MPBN \cite{guo2023membrane} & DT & ResNet-34 & 21.79 & 4                                             & 64.71                                                          \\ 
    & GAC-SNN \cite{qiu2024gated}      & DT             & MS-ResNet-34  & 21.93     & 4/6 & 69.77/70.42 \\  \midrule
    & \textbf{DTA-SNN(Ours)}       & DT             & MS-ResNet-34  & 22.02    & \begin{tabular}[c]{@{}l@{}}4\\ 6\end{tabular} & \begin{tabular}[c]{@{}l@{}}\textbf{70.27} \\ \textbf{71.29} \end{tabular} \\ 
\bottomrule                                                     
    \end{tabular}
    }}
    \caption{Comparison with SOTA methods on ImageNet-1k.}
    \label{tab:table4}
\end{table*}

\begin{table}[h]
    \centering
    \resizebox{\linewidth}{!}
    {\small{
    \begin{tabular}{@{}lccccc@{}}
    \toprule
    &Method   & NN Architecture   & T                                     & Accuracy (\%)                                                      \\ \midrule         
    & Dspike \cite{li2021differentiable}   & ResNet-18 & 10                                             & 75.4                                                             \\
    & STBP-tdBN \cite{zheng2021going}   & ResNet-19  & 4                                             & 67.8                                                             \\
    & TET \cite{deng2022temporal}   & VGGSNN & 10                                             & 77.3                                                             \\
    & TEBN \cite{duan2022temporal}   & 7-layerCNN & 10                                             & 75.1                                                  \\
    & SEW-ResNet \cite{fang2021deep}  & SEW-ResNet      & 16                                             & 74.4                                                    \\
    & MS-ResNet \cite{hu2024advancing}  & Ms-ResNet-20      & -                                             & 75.6                                                    \\
    & GLIF \cite{yao2022glif}   & 7B-wideNet & 16                                             & 78.1                                                             \\ 
    & Real Spike \cite{guo2022real}  & ResNet-20 & 10                                             & 78.0                                                            \\
    & Ternary Spike \cite{guo2024ternary}  & ResNet-20 & 10                                             & 78.7                                                            \\
    & TAB \cite{jiang2024tab}  & 7-layerCNN & 4                                             & 76.7                                                            \\
    & MPBN \cite{guo2023membrane}  & ResNet-20 & 10                                             & 78.7                                                           \\ 
    & Spikformer \cite{zhou2023spikformer}  & Spikformer-2-256 & 10                                             & 78.9                                                           \\
    & Spikingformer \cite{zhou2023spikingformerspikedrivenresiduallearning} & Spikingformer-2-256 & 10                                             & 79.9                                                           \\
    & Spike-driven \cite{yao2023spikedriven} & S-Transformer-2-256 & 16                                             & 80.0                                                           \\
    & TA-SNN \cite{yao2021temporal}    & 5-layerCNN & 10                                             & 72.0         \\
    & TCJA-SNN \cite{zhu2024tcja}   & VGGSNN & 10                                             & 80.7         \\
    \midrule
    & \textbf{DTA-SNN(Ours)} & MS-ResNet-18 & 10                                             & \textbf{81.3$\pm$ 0.3}                                                           \\
\bottomrule                                                     
    \end{tabular}
    }}
    \caption{Comparison with SOTA methods on CIFAR10-DVS.}
    \label{tab:table5} \vspace{-1em}
\end{table}

\noindent\textbf{ImageNet-1k}. 
We evaluated our DTA mechanism using the widely utilized large-scale static dataset ImageNet-1k and adopted MS-ResNet34 as the backbone, assessing the mechanism at 4 and 6 time steps, as detailed in \Cref{tab:table4}. At 4 time steps, we achieved a notable performance of 70.27{\%}. Our method, regardless of the training type, outperformed recent A2C conversion methods with substantially fewer time steps: SRP (69.43{\%}), QCFS (69.37{\%}), RMP-SNN (69.89{\%}), and DT methods: Dspike (68.19{\%}) and GLIF (69.09{\%}). At 6 time steps, we achieved an additional 0.87{\%} increase in top-1 accuracy, surpassing the best result of previous attention mechanisms \cite{qiu2024gated} at the same time step, with only a minor parameter increase of approximately 0.09. Experiments on large-scale static datasets demonstrate that the method consistently achieved strong performance regardless of the training type, approach, or number of time steps.

\noindent\textbf{CIFAR10-DVS}.
We evaluate the proposed DTA mechanism on the widely used dynamic dataset CIFAR10-DVS, which provides 0.9k training samples per label. Compared to static datasets, dynamic datasets suffer from significant noise, making them more prone to overfitting when trained with complex architectures. However, our approach, utilizing MS-ResNet-18 as the backbone, achieves best with an accuracy of 81.3{\%} at 10 time steps, surpassing previous approaches based on transformer architectures \cite{zhou2023spikformer,zhou2023spikingformerspikedrivenresiduallearning,yao2023spikedriven}. This demonstrates that our model effectively learns temporal patterns in dynamic datasets. The results of our experiments on CIFAR10-DVS are reported as the mean and standard deviation over three runs, as shown in \Cref{tab:table5}.

\begin{figure*}
  \centering
    \includegraphics[width=\textwidth]{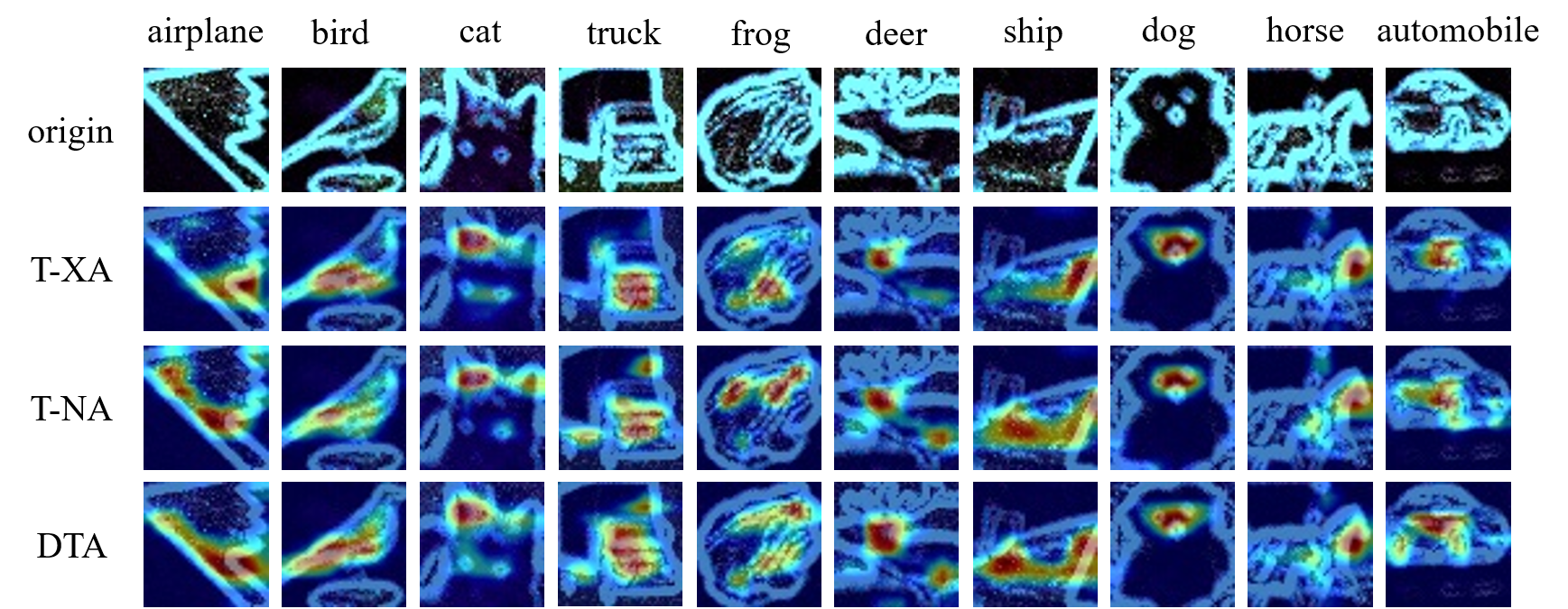}
    \caption[c]{Grad-CAM results for CIFAR10-DVS, visualized across 10 classes in four rows. The first row shows the original images, followed by Grad-CAM visualizations generated using only T-XA in the second row and only T-NA in the third row. The fourth row illustrates the results of the DTA mechanism.}
    \label{fig:figure3}
\end{figure*}

\subsection{Ablation Study} \label{subsec:ablation}
We conducted experiments on both the static CIFAR100 dataset and the dynamic CIFAR10-DVS dataset to validate the effectiveness of our proposed identical and non-identical attention approach, built upon temporal-channel-wise processing. To comprehensively assess the impact of the T-XA/T-NA modules and the DTA block, we performed a series of ablation studies. The results, detailed in \Cref{tab:table6}, demonstrate that the DTA mechanism is crucial for boosting overall performance. The DTA mechanism achieved the highest performance, with improvements of approximately 1.6{\%} and 1.2{\%} on CIFAR100 and CIFAR10-DVS, respectively, compared to the baseline. \begin{table}[h]
  \centering
  {\small{
    \begin{tabular}{ccccc}
    \toprule
    \multirow{2}{*}{T-XA} &\multirow{2}{*}{T-NA} & CIFAR100 & CIFAR10-DVS \\
     &  & (T=6) & (T=10) \\ \midrule
    {\ding{55}}  & {\ding{55}} & 79.74   & 80.4       \\
    {\ding{51}}  & {\ding{55}} & 80.56   & 80.8        \\
    {\ding{55}}  & {\ding{51}} & 80.78   & 81.2        \\
    {\ding{51}}  & {\ding{51}} & \textbf{81.28}   & \textbf{81.6}       \\ \bottomrule
    \end{tabular}
      }}
  \caption{Top-1 test accuracy ({\%}) for ablation studies of T-XA and T-NA modules.}
  \label{tab:table6}\vspace{-1em}
\end{table}This indicates that applying the attention mechanism to enrich spike representation in SNNs is reasonable. Furthermore, the inherent dynamics in SNNs, driven by temporal information, resulted in slightly more noticeable performance gains from attention mechanisms in the dynamic dataset (CIFAR10-DVS) compared to the static dataset (CIFAR100), as observed across all ablation studies. This implies that the attention mechanism, which mirrors the dynamic neural processes in humans, exerts a more pronounced effect in dynamic environments. Moreover, in most SNN architectures, the number of simulation time steps is generally higher than the number of channels, leading us to interpret this as a need to emphasize temporal-channel relationship through attention. Therefore, we considered both identical and non-identical attention mechanisms for temporal-channel correlation and dependency. As shown in the experimental results, the application of each module individually outperformed the baseline performance. Thus, the DTA combined from each module yielded the most optimal results, capturing a broader range of relevant features compared to any module used in isolation. 
In addtion, as shown in \Cref{fig:figure3}, we deployed Grad-CAM \cite{selvaraju2017grad} on 10 classes to assess the impact of removing specific components within the DTA-SNN architecture. The study highlights how our proposed attention mechanism affects the model's ability to localize key features even in dynamic data with significant noise. In summary, this demonstrates the remarkable effectiveness of the complementary dual attention operations, which accounts for both correlation and dependency within the temporal-channel domain. 

\section{Conclusion}
\label{sec:Conclusion}
In this work, we introduce a novel DTA mechanism that integrates the T-XA and the T-NA modules. Our proposed DTA mechanism addresses a gap in prior research by simultaneously exploiting both identical and non-identical attention operations to analyze temporal-channel correlation and dependency. In detail, the T-XA module focuses on temporal-channel correlation through an identical attention operation, while the T-NA module captures both local and global dependencies of the temporal-channel using non-identical attention operations. By consolidating these two strategies into a single DTA block, our proposed attention mechanism effectively enriches the spike representation and identifies variations in temporal-channel information, enhancing the comprehension of SNNs regarding complex temporal-channel relationship. Thus, extensive experiments demonstrate that our method consistently yields SOTA performance on both static and dynamic datasets, including CIFAR10 (96.73{\%}), CIFAR100 (81.16{\%}), ImageNet-1k (71.29{\%}), and CIFAR10-DVS (81.3{\%}).

\section*{Acknowledgement}
This research was funded by the STI 2030-Major Projects 2022ZD0208700, and the Fundamental Research Funds for the Central Universities.

{\small
\bibliographystyle{ieee_fullname}
\bibliography{egbib}
}

\end{document}